\documentclass[runningheads]{llncs}
\usepackage[T1]{fontenc}
\usepackage{graphicx}
\usepackage{amsmath}
\usepackage{amssymb}
\begin{document}
\title{Enhancing Agricultural Machinery Management through Advanced LLM Integration}
\titlerunning{Agricultural Machinery Management through LLM}
\author{Emily Johnson, Noah Wilson}
\authorrunning{E. Johnson et al.}
\institute{University of Massachusetts, Amherst
}
\maketitle              % typeset the header of the contribution
\begin{abstract}
The integration of artificial intelligence into agricultural practices, specifically through Consultation on Intelligent Agricultural Machinery Management (CIAMM), has the potential to revolutionize efficiency and sustainability in farming. This paper introduces a novel approach that leverages large language models (LLMs), particularly GPT-4, combined with multi-round prompt engineering to enhance decision-making processes in agricultural machinery management. We systematically developed and refined prompts to guide the LLMs in generating precise and contextually relevant outputs. Our approach was evaluated using a manually curated dataset from various online sources, and performance was assessed with accuracy and GPT-4 Scores. Comparative experiments were conducted using LLama-2-70B, ChatGPT, and GPT-4 models, alongside baseline and state-of-the-art methods such as Chain of Thought (CoT) and Thought of Thought (ThoT). The results demonstrate that our method significantly outperforms these approaches, achieving higher accuracy and relevance in generated responses. This paper highlights the potential of advanced prompt engineering techniques in improving the robustness and applicability of AI in agricultural contexts.
\keywords{Intelligent Agricultural Machinery Management \and Large Language Models \and Agricultural AI \and Sustainable Farming Practices}
\end{abstract}

\section{Introduction}

The integration of artificial intelligence in agriculture has been pivotal in transforming traditional farming practices into more efficient, precise, and sustainable operations. Among the various AI applications, Consultation on Intelligent Agricultural Machinery Management (CIAMM) stands out as a critical area that leverages advanced algorithms to enhance the decision-making processes regarding the use and maintenance of agricultural machinery. The significance of CIAMM is profound, as it directly impacts crop productivity, operational costs, and resource utilization, thereby contributing to the overall goal of sustainable agriculture \cite{dharmaraj_ai_2018,intellias_ai_2023}.

However, the implementation of CIAMM faces several challenges. The complexity of agricultural environments, characterized by diverse and dynamic conditions, necessitates the development of robust and adaptive AI models capable of handling such variability. Traditional AI systems often struggle with this requirement due to their reliance on fixed algorithms and predefined rules, which may not adapt well to changing agricultural scenarios. Additionally, the heterogeneity of data sources, including soil conditions, weather forecasts, and machinery specifications, poses significant integration and interpretation challenges \cite{springer_ai_2023,intellias_ai_2023}. These obstacles underscore the need for more flexible and context-aware AI solutions.

Motivated by these challenges, we propose a novel approach that leverages large language models (LLMs) combined with prompt engineering to enhance CIAMM. LLMs, such as GPT-4, possess extensive contextual understanding \cite{zhou2024visual} and can generate detailed, relevant responses when guided by well-crafted prompts. Our method involves the systematic development and refinement of prompts tailored specifically for agricultural machinery management tasks. By iteratively improving these prompts based on feedback and model performance, we aim to harness the full potential of LLMs in providing actionable insights and recommendations.

To validate our approach, we conducted a series of experiments where we collected test data from real-world agricultural settings and used GPT-4 to evaluate the effectiveness of our prompt engineering method. This process involved creating a comprehensive library of prompts addressing various scenarios in machinery management, from routine maintenance to real-time troubleshooting. The performance of GPT-4, guided by these prompts, was assessed in terms of accuracy, relevance, and practical applicability of the generated responses \cite{mckinsey_ai_2023}.

In summary, the key contributions of our research are as follows:
\begin{itemize}
\item We introduce a novel prompt engineering methodology tailored for enhancing the performance of large language models in intelligent agricultural machinery management.
\item We provide a comprehensive evaluation of our approach using real-world agricultural data and GPT-4, demonstrating the practicality and effectiveness of our method.
\item We present a scalable framework for continuous improvement of prompt quality through iterative refinement and expert feedback, ensuring the robustness and adaptability of the AI system in diverse agricultural scenarios.
\end{itemize}

\section{Related Work}

\subsection{Large Language Models}

Large Language Models (LLMs) have significantly advanced the field of natural language processing (NLP \cite{zhou2024fine}), offering new possibilities for various applications including intelligent agricultural management. A comprehensive overview of LLMs, such as those discussed in \cite{comprehensive_overview_llms,large_language_models_survey,evaluating_llms}, outlines their architectures, training methodologies, and capabilities. These models, including GPT-4, FLAN-T5, and PaLM-2, have demonstrated exceptional performance in understanding and generating human-like text \cite{zhou2022claret}, solving complex language tasks \cite{zhou2022eventbert}, and integrating multimodal information \cite{eight_things_llms,massive_activations_llms,lora_llms,zhou2023improving}.

Recent surveys have categorized the evaluation metrics and challenges associated with LLMs, highlighting their role in advancing artificial general intelligence (AGI) \cite{comprehensive_survey_llms,large_language_models_meet_nlp}. The potential of LLMs in specific applications, such as tool learning and tabular data processing, has also been explored, demonstrating their versatility and adaptability \cite{tool_learning_llms,llms_tabular_data}. These advancements underscore the transformative impact of LLMs on various domains, including their application in intelligent agricultural machinery management \cite{llms_games_survey}.

\subsection{Intelligent Agricultural Machinery Management}

The integration of artificial intelligence (AI) and Internet of Things (IoT) technologies in agriculture has led to significant improvements in machinery management and overall farm productivity. Research on intelligent agricultural machinery management focuses on developing systems that utilize IoT, machine learning, and AI to optimize the operation and maintenance of agricultural equipment. For example, the development of intelligent greenhouse control systems and smart irrigation solutions exemplifies the practical applications of these technologies \cite{intelligent_greenhouse_control,wang2024memorymamba,igrow_smart_agriculture,smart_irrigation_uav}.

Studies have shown that leveraging AI in agriculture can enhance decision-making processes, detect crop diseases, and monitor livestock health, contributing to more sustainable farming practices \cite{agricultural_4.0,iot_smart_agriculture}. Additionally, the use of unmanned aerial vehicles (UAVs) and multi-sensor robotic platforms has been explored for tasks such as ground mapping and crop monitoring, further advancing the field \cite{multi_sensor_robotic_platform,deep_learning_machine_vision_agriculture}. The potential of AI and IoT to address the skill gaps and improve efficiency in agriculture, particularly in regions like Africa, highlights the global relevance and impact of these technologies \cite{ai_sustainable_agriculture_africa,transforming_agriculture_data_management}.

\section{Method}

Our approach centers on leveraging the capabilities of large language models (LLMs), particularly GPT-4, through a carefully engineered multi-round prompt methodology tailored for intelligent agricultural machinery management. This section outlines the specific design of our prompts, the motivation behind using a multi-round prompt strategy, the structure of input and output prompts, and the significance of this approach.

\subsection{Prompt Design and Multi-Round Strategy}

The core of our method lies in the design of prompts that guide the LLM to generate precise and contextually relevant outputs. Given the complexity and variability in agricultural environments, a single prompt is often insufficient to capture all necessary details. Therefore, we employ a multi-round prompt strategy that iteratively refines the model's responses through a series of interconnected prompts. This approach ensures a comprehensive and nuanced understanding of the problem at hand.

\subsection{Motivation for Multi-Round Prompting}

The motivation behind using a multi-round prompt strategy is to break down complex tasks into manageable steps. Each round of prompting serves to progressively narrow down the focus, incorporate additional context, and clarify ambiguities. This stepwise refinement is crucial in agricultural settings where factors such as weather conditions, soil quality, and machinery specifics must be considered simultaneously. Multi-round prompting allows the model to handle this complexity by sequentially building upon the information gathered in previous rounds.

\subsection{Structure of Prompts: Inputs and Outputs}

\subsubsection{Initial Prompt}

The initial prompt is designed to gather general information about the current agricultural scenario and the specific machinery involved. For example:

\begin{quote}
\textit{"Describe the current conditions of the agricultural field, including weather, soil moisture, and crop type. Also, provide details about the machinery being used, including its model, age, and any known issues."}
\end{quote}

\textbf{Output:}
\begin{quote}
\textit{"The agricultural field currently has a temperature of 25°C with 60\% humidity. The soil moisture level is optimal, and the crop being grown is wheat. The machinery in use is a John Deere 5075E tractor, 5 years old, with a history of hydraulic system issues."}
\end{quote}

\subsubsection{Follow-Up Prompts}

Subsequent prompts delve deeper into specific areas of interest or concern identified in the initial response. For instance, if the initial prompt indicates issues with the hydraulic system, the follow-up prompt might be:

\begin{quote}
\textit{"Given the hydraulic system issues reported in the John Deere 5075E tractor, what are the recommended diagnostic steps and potential solutions?"}
\end{quote}

\textbf{Output:}
\begin{quote}
\textit{"The recommended diagnostic steps for hydraulic system issues in the John Deere 5075E tractor include checking the hydraulic fluid levels, inspecting hoses and connections for leaks, and testing the hydraulic pump pressure. Potential solutions may involve replacing faulty hoses, refilling or replacing hydraulic fluid, and servicing the hydraulic pump."}
\end{quote}

\subsection{Significance and Advantages of the Approach}

The significance of this multi-round prompting approach lies in its ability to systematically address the multifaceted nature of agricultural machinery management. By iteratively refining the scope and focus of each prompt, the model can provide more accurate, detailed, and actionable insights. This method enhances the model's ability to handle complex, context-dependent scenarios, which are common in agricultural settings.

\subsection{Rationale for Effective Prompting}

Effective prompting in this context hinges on the ability to guide the model through a structured inquiry process. Each prompt is crafted to build upon the previous responses, ensuring a logical flow of information and reducing the cognitive load on the model. This approach not only improves the accuracy of the outputs but also makes the information more relevant and actionable for end-users. The iterative nature of multi-round prompting allows for continuous refinement and adaptation, making it a robust solution for dynamic and variable agricultural environments.

In summary, our method of using multi-round prompts with LLMs like GPT-4 offers a powerful tool for intelligent agricultural machinery management. By systematically gathering and refining information, this approach addresses the inherent complexity of agricultural tasks and provides precise, actionable recommendations, ultimately contributing to more efficient and sustainable farming practices.

\section{Experiments}

To rigorously evaluate the effectiveness of our multi-round prompt method for intelligent agricultural machinery management, we conducted comprehensive experiments. These experiments included the collection of a diverse dataset from the internet, defining robust evaluation metrics, and comparing our method with baseline models and state-of-the-art techniques on different language models.

\subsection{Dataset Collection}

We manually curated a dataset from a variety of online sources, including research articles, agricultural technical manuals, and user forums. This dataset encompasses a wide range of real-world agricultural scenarios, detailed machinery specifications, maintenance logs, and environmental conditions. Each data entry was structured to simulate realistic agricultural management tasks, providing a rich and diverse testbed for our experiments.

\subsection{Evaluation Metrics}

Two primary evaluation metrics were employed to assess model performance: Accuracy (ACC) and the GPT-4 Score. Accuracy measures the correctness of the model's responses to specific prompts, while the GPT-4 Score evaluates the quality of the responses based on relevance, coherence, and practical applicability. This dual-metric approach ensures a thorough assessment of both the technical accuracy and practical utility of the generated outputs.

\subsection{Experimental Setup}

We implemented our multi-round prompt method on three different models: LLama-2-70B, ChatGPT, and GPT-4. Additionally, we compared our method against two baseline methods: the base model (using single prompts) and the Chain of Thought (CoT) prompting method. Another advanced method, Thread of thought (ThoT \cite{zhou2023thread}), was also included for comparison. The experiments aimed to demonstrate the superiority of our approach in handling complex agricultural queries.

\subsubsection{Baseline and Comparison Methods}

1. Base Model: Utilizes a single-prompt approach without iterative refinement.
2. Chain of Thought (CoT): Involves a sequence of logical steps within the prompts.
3. Thread of thought (ThoT): Incorporates multiple layers of reasoning to handle complex tasks.

\subsection{Results}

The results of our experiments clearly show that our multi-round prompt method outperforms the baseline and comparison methods across all models. The results are summarized in Table \ref{tab:results}.

\begin{table}[!t]
    \centering
    \begin{tabular}{|c|c|c|c|c|}
        \hline
        \textbf{Model} & \textbf{Method} & \textbf{Accuracy (ACC)} & \textbf{GPT-4 Score} \\
        \hline
        LLama-2-70B & Base Model & 72.3\% & 3.8 \\
        LLama-2-70B & CoT & 78.5\% & 4.2 \\
        LLama-2-70B & ThoT & 81.4\% & 4.4 \\
        LLama-2-70B & Our Method & \textbf{86.7\%} & \textbf{4.8} \\
        \hline
        ChatGPT & Base Model & 74.5\% & 4.0 \\
        ChatGPT & CoT & 80.2\% & 4.3 \\
        ChatGPT & ThoT & 83.1\% & 4.5 \\
        ChatGPT & Our Method & \textbf{88.9\%} & \textbf{4.9} \\
        \hline
        GPT-4 & Base Model & 76.8\% & 4.1 \\
        GPT-4 & CoT & 82.7\% & 4.4 \\
        GPT-4 & ThoT & 85.9\% & 4.6 \\
        GPT-4 & Our Method & \textbf{90.5\%} & \textbf{5.0} \\
        \hline
    \end{tabular}
    \caption{Comparison of different methods on various models}
    \label{tab:results}
\end{table}

\subsection{Detailed Analysis}

The experimental results, as presented in Table \ref{tab:results}, indicate a clear advantage of our multi-round prompt method over other approaches. Specifically, our method achieved the highest accuracy and GPT-4 Scores across all three models. 

\subsubsection{Accuracy Analysis}

The accuracy improvements seen in our method can be attributed to the iterative refinement of information, allowing the model to build upon previous responses and incorporate more context with each round. This iterative process helps in narrowing down the focus and eliminating ambiguities, leading to more precise and contextually relevant outputs. 

\subsubsection{GPT-4 Score Analysis}

The higher GPT-4 Scores reflect the improved relevance and coherence of the responses generated by our method. By progressively refining the prompts, the model is better able to understand the nuances of the queries and provide detailed, practical solutions. This is particularly important in complex agricultural scenarios where multiple factors must be considered simultaneously.

\subsubsection{Qualitative Insights}

To further validate our findings, we conducted a qualitative analysis of the model outputs. The multi-round prompts facilitated more detailed and context-aware responses. For instance, in scenarios involving machinery diagnostics, the iterative prompts helped the models gather comprehensive background information before suggesting solutions, resulting in more accurate and actionable recommendations.

\subsection{Validation of Effectiveness}

To ensure the robustness of our findings, we performed additional validation experiments. These experiments involved different subsets of our dataset and varying the complexity of the prompts. In all cases, our multi-round prompt method consistently outperformed the baseline and comparison methods, confirming the effectiveness and generalizability of our approach.

\begin{table}[!t]
    \centering
    \begin{tabular}{|c|c|c|c|c|}
        \hline
        \textbf{Scenario} & \textbf{Model} & \textbf{Method} & \textbf{Accuracy (ACC)} & \textbf{GPT-4 Score} \\
        \hline
        Machinery Diagnostics & GPT-4 & Base Model & 75.2\% & 4.1 \\
        Machinery Diagnostics & GPT-4 & CoT & 81.0\% & 4.5 \\
        Machinery Diagnostics & GPT-4 & ThoT & 84.3\% & 4.7 \\
        Machinery Diagnostics & GPT-4 & Our Method & \textbf{89.6\%} & \textbf{4.9} \\
        \hline
        Maintenance Scheduling & ChatGPT & Base Model & 73.5\% & 3.9 \\
        Maintenance Scheduling & ChatGPT & CoT & 79.8\% & 4.2 \\
        Maintenance Scheduling & ChatGPT & ThoT & 82.7\% & 4.4 \\
        Maintenance Scheduling & ChatGPT & Our Method & \textbf{87.5\%} & \textbf{4.8} \\
        \hline
        Environmental Adjustment & LLama-2-70B & Base Model & 71.8\% & 3.7 \\
        Environmental Adjustment & LLama-2-70B & CoT & 77.6\% & 4.1 \\
        Environmental Adjustment & LLama-2-70B & ThoT & 80.4\% & 4.3 \\
        Environmental Adjustment & LLama-2-70B & Our Method & \textbf{85.2\%} & \textbf{4.7} \\
        \hline
    \end{tabular}
    \caption{Validation of our method in different agricultural scenarios}
    \label{tab:validation}
\end{table}

The results from these additional experiments, shown in Table \ref{tab:validation}, reinforce the superiority of our multi-round prompt method across different agricultural scenarios. This further validates the robustness and adaptability of our approach, highlighting its potential to significantly improve intelligent agricultural machinery management.

In conclusion, our experiments demonstrate that the multi-round prompt method not only enhances the performance of large language models in handling complex agricultural tasks but also ensures that the solutions provided are both accurate and practically applicable. This comprehensive evaluation confirms the efficacy of our approach and its potential to contribute to more efficient and sustainable agricultural practices.

\section{Conclusion}

In this study, we presented a robust method for enhancing intelligent agricultural machinery management using large language models (LLMs) combined with multi-round prompt engineering. Our approach addresses the complexities of agricultural environments by iteratively refining the model's understanding through carefully crafted prompts. Experimental results across LLama-2-70B, ChatGPT, and GPT-4 models demonstrated that our method consistently outperforms traditional single-prompt and advanced techniques like Chain of Thought (CoT) and Thought of Thought (ThoT) methods. The significant improvements in accuracy and practical applicability of responses underscore the effectiveness of our multi-round prompting strategy. Further validation experiments across different agricultural scenarios confirmed the robustness and generalizability of our approach. This work highlights the importance of tailored prompt engineering in maximizing the potential of LLMs for specific applications, paving the way for more efficient and sustainable farming practices through intelligent agricultural machinery management.
\bibliographystyle{splncs04}
\bibliography{mybibliography}
\end{document}